\documentclass[10pt,twocolumn]{article}
\usepackage{times}
\usepackage{graphicx}
\usepackage{xspace}
\usepackage{epsfig,endnotes}
\usepackage{amsmath,amssymb}
\usepackage{url}
\usepackage{color}
\usepackage{soul}
\usepackage{arydshln}
\usepackage{fancybox}
\usepackage{mdframed}

\usepackage{microtype}
\usepackage{booktabs} % for professional tables
\usepackage{multirow}
\usepackage{textcomp}
\usepackage{epsfig,graphicx}
\usepackage{xcolor}
\usepackage{amsfonts,amsmath,amssymb}
\usepackage{booktabs}
\usepackage{amsmath,amssymb,amsfonts}
\usepackage{algorithmic}
\usepackage{graphicx}
\usepackage{bm}
\usepackage{comment}
\graphicspath{{imgs/}}
\usepackage[utf8]{inputenc}
\usepackage{float}
\usepackage{threeparttable}
\usepackage{subcaption}
\usepackage{url}
\usepackage[colorlinks=true,citecolor=magenta,linkcolor=magenta,urlcolor=magenta]{hyperref}
\usepackage{balance}

\renewcommand{\vec}[1]{\bm{#1}}
\newcommand{\E}[1]{\mathop{\mathbb{E}[{#1}]}}
\newcommand{\bfitDelta}{\bm{\mathit{\Delta}}}
\DeclareMathOperator*{\argmax}{arg\,max}

\newcommand{\sysname}{Griffin\xspace}
\newcommand{\companyname}{Microsoft\xspace}

\definecolor{lightgray}{rgb}{.8,.8,.8}  % define new color
\newcommand{\mypara}[1]{\vspace{2mm}\noindent\textbf{#1}}

\newcommand{\estim}[1]{${#1}^\beta$}

\begin{document}

\title{\sysname: Reasoning about Job Anomalies with\\ Unlabeled Data in Cloud-based Platforms}
\author{
Liqun Shao$^1$, Yiwen Zhu$^1$, Abhiram Eswaran$^1$, Kristin Lieber$^1$, \\
	Janhavi Mahajan$^1$, Minsoo Thigpen$^1$, Sudhir Darbha$^1$, \\ Siqi Liu$^2$, Subru Krishnan$^1$, Soundar Srinivasan$^1$, \\
	Carlo Curino$^1$ and Konstantinos Karanasos$^1$ \\
\small {\em $^1$Microsoft Corp. \quad
	$^2$University of Pittsburgh, PA, USA} \\ [2mm]
%\small Submission Type: Experience
}
%\author{John Doe$^1$ and Jane Monroe$^2$ \\
%\small {\em  $^1$Dept. Cloud, Cloud University \quad
%          $^2$Cloud National Labs} \\ [2mm]
%\small Submission Type: Experience
%}
\date{}
\maketitle

\begin{abstract}
	
	\companyname's internal big data analytics platform is comprised of hundreds of thousands of machines, serving over half a million jobs daily, from thousands of users. 
	The majority of these jobs are recurring and are crucial for the company's operation. 
	Although administrators spend significant effort tuning system performance, some jobs inevitably experience slowdowns, i.e., their execution time degrades over previous runs.
	Currently, the investigation of such slowdowns is a labor-intensive and error-prone process, which costs \companyname significant human and machine resources, and negatively impacts several lines of businesses.
	% , involving complicated analysis and error triaging.
	% This is , and leads to long wait times between problem reporting, assessment, and resolution, which 
	% This costs Microsoft significant human and machine resources, and negatively impacts several lines of businesses.
	
	In this work, we present \sysname, a system we built and have deployed in production last year to automatically discover the root cause of job slowdowns.
	Existing solutions either rely on labeled data (i.e., resolved incidents with labeled reasons for job slowdowns), which is in most cases non-existent or non-trivial to acquire, or on time-series analysis of individual metrics that do not target specific jobs holistically.
	In contrast, in \sysname we cast the problem to a corresponding regression one that predicts the runtime of a job, and show how the relative contributions of the features used to train our interpretable model can be exploited to rank the potential causes of job slowdowns.
	Evaluated over historical incidents, we show that \sysname discovers slowdown causes that are consistent with the ones validated by domain-expert engineers, in a fraction of the time required by them.

\end{abstract}

\section{Introduction}

\companyname operates one of the biggest data lakes worldwide for its big data analytics needs~\cite{hydra}. It is comprised of several clusters for a total of over 250k machines and receives approximately half a million jobs daily that process exabytes of data on behalf of thousands of users across the organization. The majority of these jobs are recurring and several of them are critical services for the company.
Hence, administrators and users put significant effort in tuning the system and the jobs to optimize their performance.
Nevertheless, some jobs inevitably experience slowdowns in their execution time (i.e., they take longer to complete than their previous occurrences) due to either system-induced (e.g., upgrades in the execution environment, network issues, hotspots in the cluster) or user-induced reasons (e.g., changes in job scripts, increase in data consumed).

% such as Azure billing, Bing services, Bing ads pipelines, Office analytics, and the Windows telemetry pipelines. 

Such \emph{job slowdowns} can have a catastrophic impact to the company. 
In fact, runtime predictability is often considered more important than pure job performance in recurring production jobs~\cite{morpheus}.
First, several jobs are interdependent, that is, the output of a job might be consumed by multiple other jobs~\cite{owl}. 
Thus, the slowdown of the first job can have a cascading effect on all other dependent jobs, impacting vital services across the company.
Second, some business-critical jobs are associated with deadlines in the form of service-level objectives (SLOs). Missing those SLOs can result in substantial financial penalties in the order of millions of dollars.

%--------------
%\begin{figure*}
%\centering
%\includegraphics[scale=.6]{DRI.PNG}
%\caption{Current and revised workflow for DRIs with Anomaly Reasoning} \label{fig:dri}
%\end{figure*}
%--------------

Despite the importance of promptly resolving such incidents, the current approach remains largely manual.
Job slowdowns are signaled either through tickets raised by customers or by missed deadlines (for jobs with SLOs).
In either case, a slow, labor-intensive process of error triaging and root-cause analysis must be initiated. 
In particular, on-call engineers 
% referred to as Direct Responsible Individuals (DRIs), 
manually investigate causes of job slowdowns by analyzing hundreds of logs and system traces through a complex monitoring dashboard.
%This process
%, depicted in the upper part of Figure~\ref{fig:dri}, 
% takes one to two hours daily, and i
Despite the existence of detailed metrics, it can sometimes take several hours to resolve an incident.
% \LS{In a month alone, millions of CPU hours are wasted due to incidents that remained unresolved, despite the existence of detailed metrics.}
This bottleneck costs millions of dollars in engineering time wasted on investigation and in job SLO violations, and results in degraded user experience. 

% In this work, we demonstrate how this process can be automated through the use of machine learning.
In this work, we present \sysname, the system we built and have deployed in our production big data analytics infrastructure to automatically discover the main factors causing a job's runtime deviation through the use of machine learning.
% Cosmos, without the need for labeled data. 
% \sysname automatically discovers , which has several benefits.
% Having a way to automatically discover the main factors causing a job's runtime deviation would greatly improve the above situation.
\sysname greatly improves the situation described above.
First, it helps users find user-induced causes of their job slowdowns and prevents them from raising tickets that are ``false alarms'' to system administrators.
Second, in case of actual infrastructure issues, it directs administrators towards the most probable causes for a job slowdown and allows
early elimination of factors unrelated to the slowdown.
Third, by observing slowdowns in jobs submitted for testing purposes, administrators can resolve system issues before they affect user jobs.

Existing related works have used detection methods such as classification and clustering to perform analysis of anomalies in cloud computing~\cite{agrawal2015survey,modi2013survey}.
However, to analyze anomalies, 
% \YZ{"to identify the job slowdown reasons"?, to be weaker as some of those methods are not strictly for job slowdown but general anomaly reasoning/detection }
these methods rely on labeled data, e.g., data from existing incidents that associate jobs with their slowdown causes. 
Such labeled training data in production cloud systems are extremely hard to obtain and can also be erroneous.
A few approaches do consider unlabeled data, but rely either on time-series analysis or restrict their focus to machine or VM behavior~\cite{cherkasova2009automated,cohen2004correlating,dean2012ubl,gu2009online,tan2010adaptive,tan2012prepare,zhang2013intelligent}.
In contrast, we focus on job instances that span several hundreds of machines but only during the lifetime of the job.
% As a result, it is not feasible to apply existing techniques for identifying root causes of job slowdowns.
As a result, existing techniques are frequently not applicable to identify root causes of job slowdowns.

Unlike existing works, \sysname employs an interpretable regression model to predict job runtime and then suggest reasons for runtime deviations.  \sysname exploits two  characteristics of available data.  First, in our clusters we collect valuable telemetry at various levels of abstraction (at the job, machine, and cluster level).  Second, the majority of our jobs are recurrent, meaning similar jobs which run regularly, for example every day or several times a day, allowing \sysname to leverage this historic data. Then, based on the relative contribution of each metric/feature to the runtime of a job that experienced a slowdown, we emit a list of possible causes for the slowdown, ranked by their importance.

% In this paper, we present \sysname, the system we built and have deployed in production to reason about job slowdowns in Cosmos, without the need for labeled data. 
% The revised workflow followed by \sysname for investigating job slowdowns is depicted in Figure~\ref{fig:DRI2} described in Section \ref{sec:problem}.

% add line space
Our contributions are the following:
\begin{enumerate}
	\item We present an end-to-end ranking system to identify the root causes of job slowdowns without human-labeled data.
	\item We show how an interpretable regression model can be used to reason about job slowdowns.
	\item We experimentally compare various models in terms of accuracy, scalability in the model size and number of jobs, and generalizability to jobs not seen before by the system.
	% show how the former can be efficiently used for jobs that \sysname has not been trained on.
	\item \sysname is deployed in our clusters and is used by our engineers. Early indications show that slowdown causes generated by \sysname are closely correlated to causes validated by domain experts. At the same time, \sysname drops the time of investigation by orders of magnitude compared to the existing manual process.
\end{enumerate}

The rest of the paper is organized as follows. Section~\ref{sec:background} provides details on our production environment and the relevance of the problem we focus on.
Section~\ref{sec:architecture} gives an overview of \sysname.
Section~\ref{sec:reasoning} describes our anomaly reasoning algorithm.
Section~\ref{sec:data collection} discusses feature engineering and data collection, whereas Section \ref{sec:deployment} provides details on \sysname's deployment.
Section~\ref{sec:experiment} presents the results of our experimental evaluation. 
Section~\ref{sec:related} discusses related work, and Section \ref{sec:conclusion} provides our concluding remarks.

\section{Background on our Environment}\label{sec:background}

In this section, we provide background on the characteristics of our analytics clusters to give a sense of the scale of the problem we target (Section~\ref{sec:bg:env}).
Then, we describe the current state of affairs in finding the reasons for a job's slowdown (Section~\ref{sec:bg:debug}).

%-------------------------------------------
\subsection{Cluster Characteristics}
\label{sec:bg:env}

At \companyname we operate a massive data infrastructure, powering our internal analytics processing.
This infrastructure consists of several clusters, each comprised of tens of thousands of machines---Table~\ref{tab:req3} highlights some details of our clusters' scale.
To make the situation even more challenging, our cluster environments are also heterogeneous, including several generations of machines.

\begin{figure}[t]
	\centering\includegraphics[width=\columnwidth]{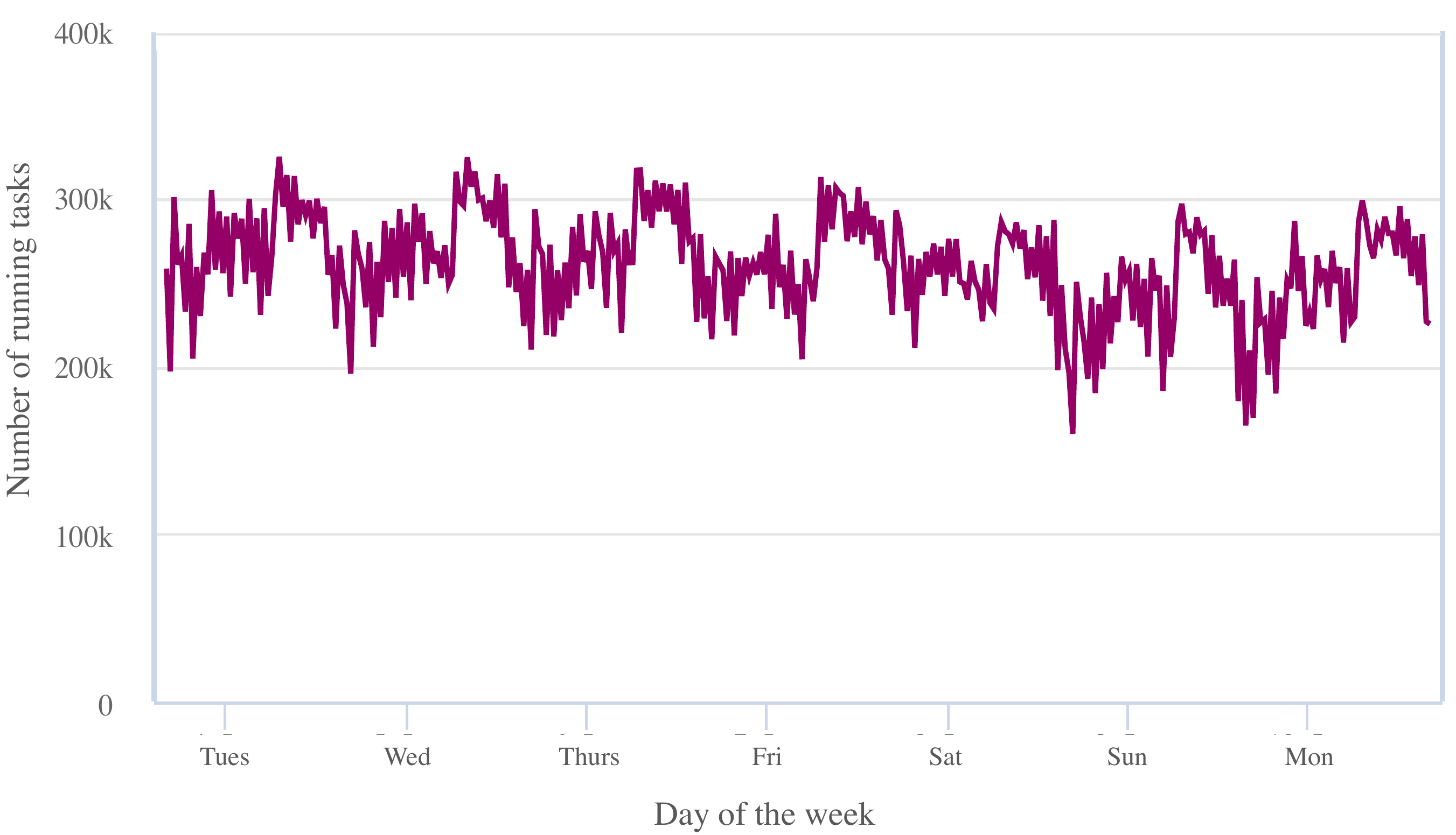}
	% \vspace{-1cm}
	\caption{Running tasks in one of \companyname's production analytics clusters, comprised of tens of thousands of machines.\label{fig:tasks}}
\end{figure}

\begin{table}[t]
	\center
	\scalebox{0.93}{
		\begin{tabular}{|l|l|l|}
			\hline 
			{\bf Dimension} & {\bf Description} & {\bf Size} \\ 
			\hline 
			\hline 
			Daily Data I/O & Total bytes processed daily  & $>$1EB \\ 
			\hline  
			Fleet Size & Number of servers in the fleet & $>$250k \\ 
			\hline  
			Cluster Size & Number of servers per cluster & $>$50k \\ 
			\hline  
		\end{tabular}
	}
	\caption{\companyname cluster environments. 	\label{tab:req3}}
	\vspace{-2mm}
\end{table}

\begin{figure*}[t]
	\centering\includegraphics[width=0.99\textwidth]{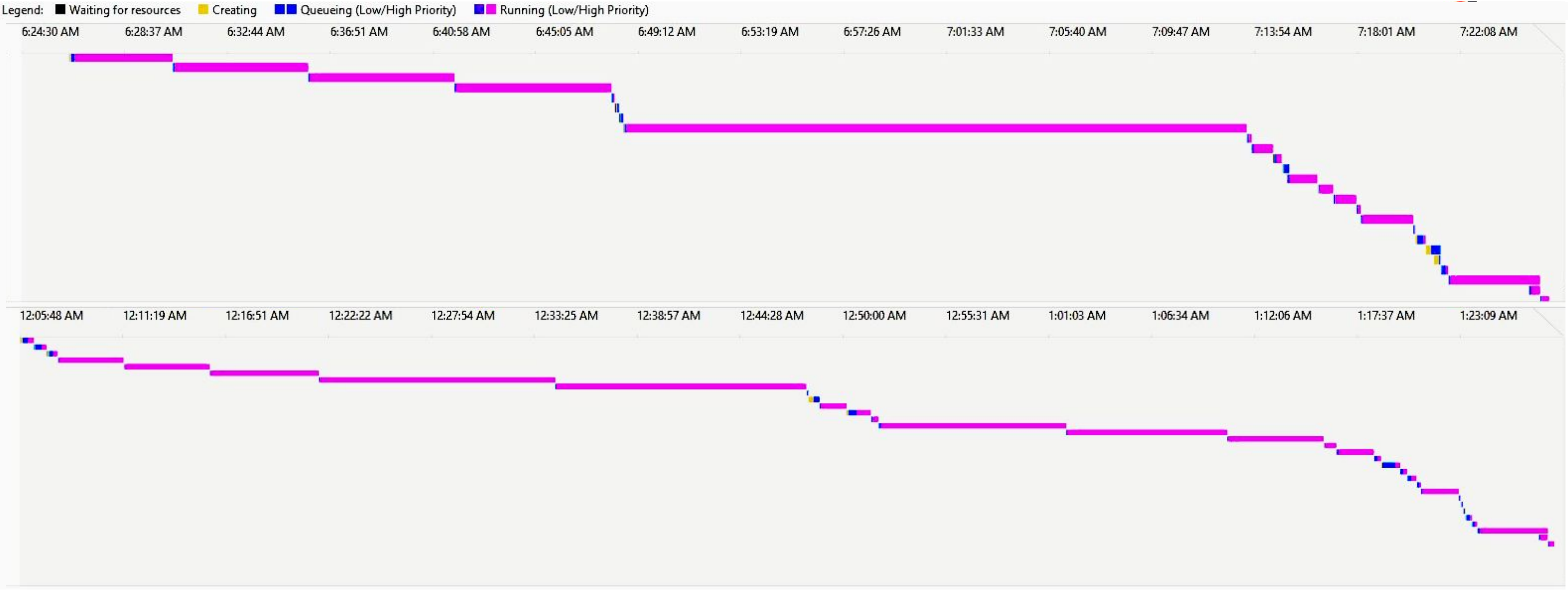}
	% \vspace{-1cm}
	\caption{Two occurrences of the same job, broken down . The bottom one takes 90\% longer than the top one.}
	\label{fig:manual}
\end{figure*}

Tens of thousands of users submit hundreds of thousands of jobs to these clusters daily. 
Each job is a directed-acyclic graph (DAG) of operators (which we term \emph{stages}), and each stage consists of several tasks~\cite{scope}. 
Each task gets executed in a cluster's machine (and each machine runs several tasks in parallel).
Figure~\ref{fig:tasks} depicts the number of running tasks in one of our clusters over the course of a week. At each moment in time there are between 200k--300k tasks running.

Given this extreme scale and complexity, job slowdowns are quite common. Manually investigating such slowdowns, as we explain in the following section is a painful and time-consuming effort.

% Our target cluster environments are very large in scale and heterogeneous, including several generations of machines.

%-------------------------------------------
\subsection{Manual Job Slow-down Investigation}
\label{sec:bg:debug}

We now describe how \companyname engineers used to approach job slowdowns before \sysname got deployed in our clusters.
Figure~\ref{fig:manual} shows two occurrences of the same job. The top one corresponds to its regular execution, taking 44~mins to complete.
The bottom one experiences a slow down with a completion time of 88~mins. The figure visualizes how long the various stages of the job take to execute 
(although several stages might run in parallel, this tool shows the ones in the critical path, as those determine the job's execution time).

% In the analytic platform, each job consists of multiple stages which are executed sequentially. During the execution, a critical path includes all the stages that take the longest time to finish counting from the starting of this job given the dependency of the stages. Any slowdown of those stages will result in a longer job execution time.

% Figure~\ref{fig:debug} shows the starting and ending time of all the critical stages of two occurrences of the same job. A typical job runs in 48 min (top figure) while the second job takes 83 min. For a DRI, the investigation of the job slow involves pulling data

To investigate this slowdown, an engineer will typically start by looking at the visualization tool of Figure~\ref{fig:manual}, trying to detect the stages that seem abnormal. Note, however, that in this example, the ``regular'' top occurrence is the one that seems to have longer stages.
Therefore, this tool is of limited use.
Next, the engineer will have to manually combine several other tools and system files to get more information about the job and the system during the time this job was executed. Given the scale of the system and the amount of metrics collected, this process can take a considerable amount of time to complete.
Multiply this by the number of slowdowns and one can easily see the significant opportunity in saving engineering time and improving user experience by speeding up this process.
Note also that only a few engineers have the knowledge to perform this manual analysis.

% We would have to manually combine different sources from various tools of the system and files. Log into machines. Do the analysis manually.
% Griffon automates this.
% It can get smarter as we go.
% All these slowdowns can be performed only by a few people in the company.
% We get to a conclusion much faster.

% \begin{figure*}
% \centering
% 	\includegraphics[width=0.8\textwidth]{job.pdf}
% \caption{Execution of a Typical Job} \label{fig:job}
% \end{figure*}

% {\color{red}Figure~\ref{fig:job} shows an example of the execution of a job, consisting of 6 stages before generating the output. Each stage, e.g. SV1\_Extract, SV2\_Aggregate, etc., can be executed with various degrees of parallelism thus runs on multiple containers across many machines. To complete a job, there can be often over thousands of machines across different locations to be involved. If anyone of the machines runs into a problem, the job may fail or slow down. Between stages and computation threads, there can be issues with networking, data skew, storage, etc. Within each thread, the queuing, deployment/scheduling of the containers might raise red flags as well. Within each machine, as multiple containers from all different jobs are executing simultaneously, the competition of CPU processors may also slow jobs down. 
% }

\section{System Overview}\label{sec:architecture}

\begin{figure*}
	\centering
	\includegraphics[width=0.85\textwidth]{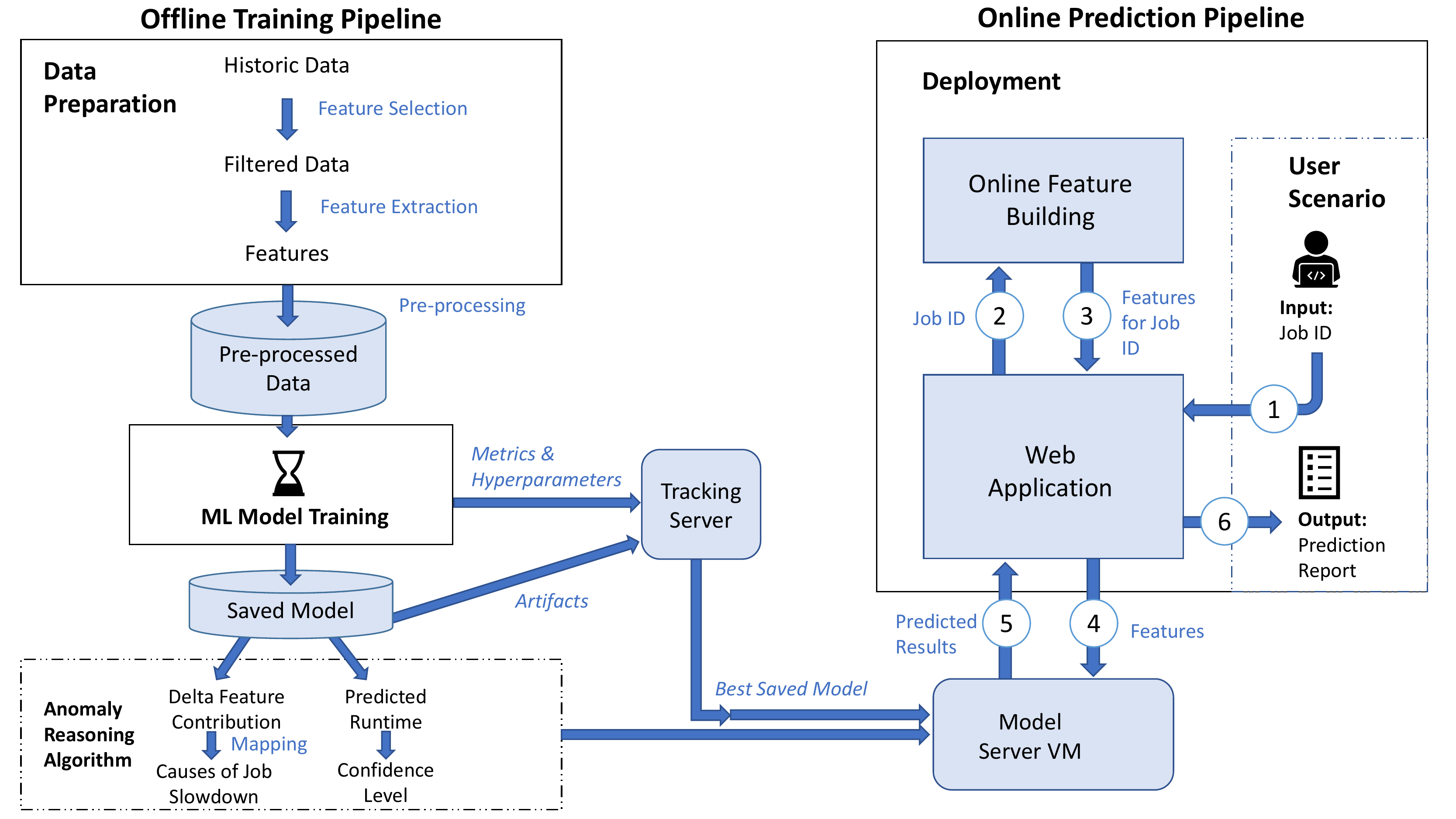}
	\caption{System architecture.} \label{fig:frm}
\end{figure*}

% \KK{Talk a bit more about absence of labeled data.}

\sysname's goal is to find the causes for job runtime degradations in our big data analytics clusters. A central requirement is to not rely on labeled data, i.e., there should be no need for existing slowdown instances associated with their causes.
% \KL{rely on labeled data for training the model.....}
Each job in our clusters is associated with a set of telemetry data that we already collect for monitoring and debugging purposes (e.g., number of tasks, size of input data, load of machines the job was executed on---see Section~\ref{sec:data collection} for details), some of which contributed to the job's slowdown.
Instead of finding a subset of the slowdown causes (as a system that relies on labeled data would do), \sysname \emph{ranks} the causes (i.e., the features) in the order they affected the deviation of the job's runtime from its expected runtime, and then suggests the top causes to the users.
A formal description of the problem and our approach for solving it is presented in Section~\ref{sec:reasoning}.

% \footnote{As we show, \sysname is capable of informing the user when the used features are not sufficient to explain the slowdown---note that this is a quite uncommon scenario, given the features we currently use in production.}

\sysname's architecture is depicted in Figure~\ref{fig:frm}.
It consists of two pipelines: the offline training and the online prediction, which we detail below.

\mypara{Training} The offline training process uses various metrics that we collect at the job, machine, and cluster level to generate a model that will be able to predict the runtime of a job given these features. Our anomaly reasoning algorithm will use this model to rank the features that contributed to the job's slowdown. The training process involves the following steps: 
(1)~data preparation, i.e., collect and clean the data from different sources in the cluster at regular intervals; 
(2)~feature engineering, i.e., extract raw features from the collected data, create new ones, and choose the ones that we will be using to train the model;
and (3)~model generation, i.e., the creation of the model, including hyper-parameter tuning, training, and model evaluation. The data preparation is detailed in Section~\ref{sec:data collection}, whereas the model details are in Section~\ref{sec:reasoning}.

% downloads the data and trains the machine learning model. It is triggered at regular intervals. 
% \begin{itemize}
% 		\item Model training: Creation of the model, including hyper-parameter tuning, training, and model evaluation.
% 		\item Data Preparation: Collecting data from different sources and download it to a local training environment regularly.
% 		\item Preprocessing: A sequence of steps that cleans the data, engineers features and filters data as shown in Section \ref{sec:data collection}.		
% \end{itemize}

\begin{figure}[t]
	\centering\includegraphics[width=0.9\columnwidth]{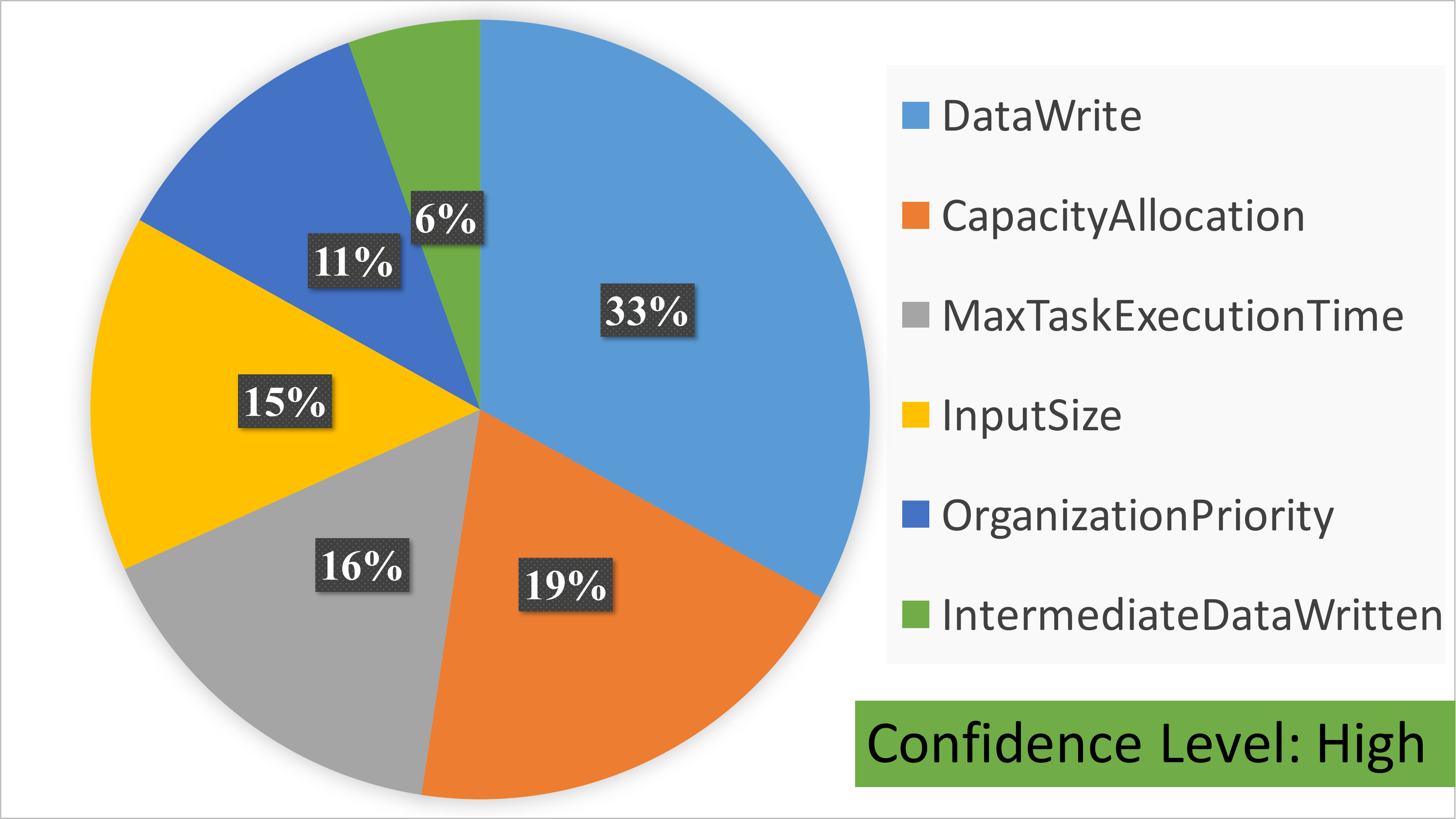}
	% \vspace{-1cm}
	\caption{\sysname's output for the slowdown of the job depicted in Figure~\ref{fig:manual}.}
	\label{fig:griffon-output}
\end{figure}

The generated model is stored in the Tracking Server, which tracks model runs and stores performance metrics and hyper-parameter values (see Section~\ref{sec:deployment}).

\mypara{Prediction} The online prediction pipeline provides an API that takes as input the ID of a job that was executed and experienced a slowdown. 
This API is exposed to the users through a web application.
% \KL{This API is exposed through a web application. (don't think you need to introduce the concept of cluster users)}
Then the Online Feature Building component gathers the metrics associated with that job and provides the data to the Model Server where the prediction model is deployed.  The output is a report with ranked reasons for job slowdown. In addition, the system provides a confidence level in the results (see Section~\ref{sec:confidence}).

High confidence means that users can rely on the output of the system. On the other hand, low confidence means that the metrics used are not sufficient to explain the job slowdown.  The output of the system is useful even in the latter case, because users can rule out these metrics and focus their investigation in other areas.

Figure~\ref{fig:griffon-output} shows \sysname's output for the job slowdown of Figure~\ref{fig:manual}. In this case, \sysname suggests with high confidence that the increase in data written by the job is the main reason for its slowdown. As this is a user-induced reason and not a problem with the system, the corresponding ticket can be closed without further investigation.

\section{Anomaly Reasoning Algorithm} \label{sec:reasoning}

In this section, we describe our algorithm for reasoning about anomalies. 
First, we formally define the problem (Section~\ref{sec:propstate}) and discuss model interpretability (Section~\ref{sec:model}). Then we describe the interpretable tree-based model that \sysname uses for determining the reasons for a job's slowdown (Section~\ref{sec:treemodel}) and its associated confidence level (Section~\ref{sec:confidence}).

\subsection{Problem Statement}\label{sec:propstate}

We consider a set of jobs that have already been executed. Hence, we know the runtime of each job. Through our collected metrics, we also know the values of the features that we are interested in (see Section~\ref{exploreselection} for feature selection).

The majority of the jobs submitted in our clusters are analytics jobs\footnote{Analytics jobs are executed using Scope, an internal SQL-like distributed query engine that enables processing of petabytes of data per job~\cite{scope}.} that are recurring, i.e., they are submitted at regular intervals (typically hourly, daily, or weekly)~\cite{hydra}. We use the notion of \textbf{job template} to refer to each of these recurring jobs. Jobs belonging to the same template have very similar scripts with minor differences, e.g., to access the latest data. 

% \KL{do you need to say Scope here? how about generalizing to "have very similar programming scripts with minor difference, e.g., using the latest data.} 

\begin{figure} [t]
	\centering
	\includegraphics[width=0.8\columnwidth]{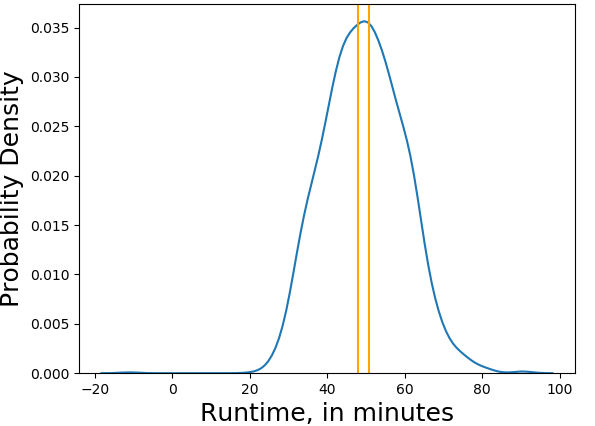}
	\caption{Example of baseline selection for a job template.} \label{fig:baselines}
\end{figure}

We also define the \textbf{baseline} of a job to be its ``expected'' runtime, given the runtime of the other jobs that belong to the same template. 
In practice, we use the mean runtime of the jobs whose runtime falls between the 45$^{th}$ and 55$^{th}$ percentile for that template.
A benefit of using a percentile measure is that we avoid outliers. Therefore ``slow jobs" in our training data will not affect the baseline set. Figure~\ref{fig:baselines} shows the runtime distribution for one job template. The data we use for baseline selection falls between the two orange lines.

For jobs that belong to templates with no previous occurrences, we use the baseline of jobs with similar characteristics (in data size and performed operations).
Similarly, we define the baseline of various features of a job to be their expected value, given the jobs of the corresponding template.

% \LS{how to get the baseline with similar characteristics? in our case, we just use feature contribution}

% This is the runtime against which we will compare a job's runtime to determine how much it has deviated from its expected behavior. 

% \KK{Below para now mostly redundant with Section 2 -- will fix.}
% In this paper, instead of trying to find a subset of the true causes, we formulate the problem as a ranking task using unlabeled data. We rank the causes by their influences on the job runtime deviation, and then suggest the top suspected causes to the users.
% For each job we have a list of real numbered features, measuring different aspects of the job, that may influence its runtime. An example feature could be input data size, or other computed statistics of the job. 

Let $\vec{x} \in \mathbb{R}^p$ be the $p$-dimensional features, and $y \in \mathbb{R}$ be the job runtime.
Let \estim{y} and \estim{\vec{x}} be the baseline of the runtime and the features, respectively. 
We define the problem as follows: for each job $i$, lacking human labeled reasons, with features $\vec{x}^i$ and runtime $y^i$, predict the rank of different features based on their influence on the deviation of $y^i$ from \estim{y^i}.

%Without more information about the job, we would expect $y$ to be close to $\E{y}$.
% \footnote{We observed from data that the original job runtime follows a log-normal distribution, so $y$ follows a normal distribution.}
% \LS{I don't think we did log-transformation.}

%a set of numbers $j_1, j_2, \ldots, j_p$, such that $x^i_{j_1}, x^i_{j_2}, \ldots, x^i_{j_p}$ are ordered by their influence on the deviation of $y^i$ from $\E{y}$. Therefore, $j_1$ is the most important feature contributing to the deviation of $y_i$ from $\E{y}$ followed by $j_2$ and so on. 

\subsection{Interpretable Model}\label{sec:model}

Consider a machine learning model that is trained to predict the runtime $y$ of a job using a set of features $\vec{x}$. That alone would be a standard regression problem. 
However, in our setting we want to use such a runtime prediction to find the features that contribute the most to a job's slowdown, that is, to the runtime's deviation from the job's baseline \estim{y}.
To this end, we need an \emph{interpretable} regression model for the job's runtime.

We define a regression model to be interpretable, if the output of the model can be expressed as the sum of contributions of each of the model's features:

\begin{equation} \label{eq:linear_contrib}
y = c + \sum_{k=1}^K fc_{k}
\end{equation}

\noindent where $c$ is a constant and $fc_{k}$ is the contribution of feature $x_k$ to the prediction. 
{\color{blue}Similarly, for a baseline job, let $y'^\beta$ be the predicted runtime based on the same model using the baseline features $x_k^\beta$, we can decompose the model prediction as:
	\begin{equation} \label{eq:linear_contrib_baseline}
	y'^{\beta} = c + \sum_{k=1}^K fc_{k}^\beta
	\end{equation}
	\noindent where $fc_{k}^\beta$ is the contribution of feature $x_k^\beta$ to the prediction. 
	
	In our setting, 
	%having an interpretable model to predict job runtime $y$ and comparing it with the baseline prediction $y'^\beta$, 
	we can quantify the \textbf{delta feature contribution} $\bfitDelta fc_{k}$ of each feature $x_k$ to the deviation of $y$ from $y'^\beta$:
	
	\begin{equation} \label{eq:delta_contrib0}
	y - y'^\beta = \sum_{k=1}^K (fc_{k}-fc_{k}^\beta) = \sum_{k=1}^K \bfitDelta fc_{k}
	\end{equation}
	
	In our case, for the baseline jobs of all templates, the model prediction is very accurate (with Mean Absolute Ratio Error as 2.2\%). Thus the sum of the delta feature contribution approximately equals to the deviation from the baseline job runtime $y^\beta$. If the predicted baseline runtime is not accurate, i.e. the difference between $y^\beta$ and $y'^\beta$ is large, we can raise a flag about our confidence of the model result, as will be discussed in Section~\ref{sec:confidence}.
}

Being able to quantify the contributions of each feature to a job's slowdown allows us to rank the features in order of importance, which is the goal of \sysname.

\mypara{Model choice}
We considered various model categories to predict the runtime of a job, namely Linear Regression (LR), Random Forest (RF), Gradient Boosted Trees (GBT), and Deep Neural Network (DNN).
Our main requirements were that the model be interpretable and that it offers good accuracy.

A linear model can be expressed as $y = \alpha + \sum_{k=1}^K \beta_k x_k$, where $\alpha, \beta_k \in \mathbb{R}$. It is trivial to show that it satisfies the interpretability criterion of Eq.~\ref{eq:linear_contrib}. However, as we show in Section~\ref{choosemodel}, the accuracy is worse than that of the other models.
GBT and DNN have acceptable accuracy but their interpretability is hard to establish.
Lastly, the RF model exhibited the best accuracy in our experiments and therefore, is our model of choice in \sysname. In the next section, we describe an appropriate tree interpreter that reformulates a tree-based model to a linear form, so that we can use it to rank feature contributions to job slowdowns.

% While the RF model has the best accuracy, it is hard to transfer a tree-based model into a linear representation as in Eq.~\ref{eq:contrib}. In the next section, we describe an appropriate tree interpreter that reformulate the tree model in a linear form.
% For easier interpretation when comparing across job instances, after we get the contributions from the features and the noise $\epsilon$, we normalize them by the sum of their absolute values.

Note that when training our models, we considered both a \emph{global} and \emph{per-template} models. In the former case, we train a single unified model to predict runtime using jobs of all templates together in the training set. In the latter, we train one model per job template utilizing training data drawn exclusively from jobs that belong to the particular template. In Section~\ref{sec:experiment}, we compare the two approaches in terms of accuracy, scalability, and generalizability.

\subsection{Interpretable Random Forest} \label{sec:treemodel}

In a Random Forest (RF) model, for each tree, in order to make a prediction, we traverse a path from the root of the tree to a leaf.
This path consists of a series of decisions based on the model's features. 
Assuming there are $M$ nodes on the path, each node separates the feature space into two, given a feature $x_k$ and a threshold $t_k$: the one child node corresponds to $x_k \leq t_k$, the other to $x_k > t_k$. 
In other words, from the root node where all the samples reside, a partition based on feature $x_k$ and threshold $t_k$ thus separates the data samples to the two children that correspond to smaller feature spaces.

Consider a tree $j$ of the model and a node $m \in j$ that is partitioned from its sibling based on feature $x_k$.
Let $\bar{y}_{m,j}$ be the mean target value for all samples that reside on node $m$. Then the contribution of feature $x_k$ to the final prediction due to this partitioning is calculated as:
\begin{eqnarray}\label{eq:tree1}
\Delta_{m}\text{contrib}_j(x,k)=(\bar{y}_{m,j} - \bar{y}_{m-1,j})I_j(m,k)
\end{eqnarray}

\noindent for $2<m\leq M$, where node $m$$-$$1$ is $m$'s parent. 
$I_j(m,k)$ equals to 1 if the partitioning at node $m$$-$$1$ involves feature $x_k$ for tree $j$ or 0 otherwise. 
The number of samples that reside on each node becomes smaller and smaller by traversing the path, as the feature space gets smaller.
The contribution of $x_k$ to the final prediction can be calculated as the sum of all $\Delta_{m}\text{contrib}_j(x,k)$:

\begin{eqnarray}\label{eq:tree2}
\text{contrib}_j(x,k)=\sum_{m=2}^{M}\Delta_{m}\text{contrib}_j(x,k) \\ =\sum_{m=2}^{M}(\bar{y}_{m,j} - \bar{y}_{m-1,j})I_j(m,k)
% &=&
\end{eqnarray}

The prediction of the target value from this tree is $\bar{y}_{M,j}$ and can be expressed using the sum of all features' contributions along the path:
\begin{eqnarray}\label{eq:tree3}
y_j = \bar{y}_{M,j} = \bar{y}_{1,j} + \sum_{m=2}^{M}\bar{y}_{m,j} - \bar{y}_{m-1,j} \\ = c_{j} + \sum_{m=2}^{M}\sum_{k=1}^{K}\Delta_{m}\text{contrib}_j(x,k)
\end{eqnarray}

\noindent where $c_{j}$ is the full sample mean.
%
%
%Each node corresponds to a feature space $R_m$ and along with the path, this region becomes smaller and smaller. Denote the feature of its parent node for the separation as $x_k$. Define the mean target value for all the samples in the features spaces for the parent node $m$ as $\bar{y}_m$
%
%
% 
\textit{TreeInterpreter}~\cite{TreeInterpreter} combines the results of all trees in our Random Forest by taking the sum of the contribution from each tree. 
Thus, each prediction is decomposed into a sum of contributions from the features, as follows:
% There are two ways to get feature importance: Scikit-Learn or TreeInterpretor.
%\textit{TreeInterpreter} can provide positive and negative feature contributions of tree-based models \cite{positivetree}. The idea itself is derived from decision trees before it is expanded to random forest.
%can be generalized to other methods related to decision trees, although perhaps a few adjustments needed to be made. 
%This is possible because the idea itself is derived from decision trees before it is expanded to random forest. 
\begin{equation} \label{eq:tree_contrib}
y = \dfrac{1}{J}\sum_{j=1}^J c_{j} + \sum_{k=1}^K (\dfrac{1}{J}\sum_{j=1}^J \text{contrib}_j(x, k))
\end{equation}
\noindent where $J$ is the number of trees, $c_{j}$ is the full sample mean for each $j^{th}$ tree, and $K$ is the number of features involved.
% , and $\text{contrib}_j(x, k)$ is the contribution from the $k^{th}$ feature in the feature vector $x$ in tree $j$.
%The form is similar to linear regression in Equation \ref{eq:linear}. For linear regression the coefficients $\beta$ are fixed, with a single constant $\epsilon$ that denotes learned bias. For the decision tree, the contribution of each feature is not a single predetermined value, but depends on the rest of the feature vector which determines the decision path that traverses the tree and thus the contributions that are passed along the way.

Using $c = \dfrac{1}{J}\sum_{j=1}^J c_{j}$ for the average runtime across the whole training set and $fc_{k} = \dfrac{1}{J}\sum_{j=1}^J \text{contrib}_j(x, k)$ for the contribution of feature $x_k$ to the predicted runtime, we get to Eq.~\ref{eq:delta_contrib0}, which shows that our Random Forest model meets the interpretability criterion. Therefore, it can be used to detect reasons for job slowdowns in \sysname.

\subsection{Confidence Level}\label{sec:confidence}

The confidence level shows how reliable is the prediction made by our model for the contribution of each feature to a job's slowdown.
We consider two factors that affect our model confidence: (1)~the relative error in predicting the runtime of the job (by comparing the model prediction with the actual runtime of the job); (2)~the confidence intervals estimated by the random forest~\cite{Mein2006}. 

The relative error is defined as following:
\begin{equation} \label{eq:error}
\text{error\_rate} = \frac{|\text{predicted\_runtime} - \text{actual\_runtime}|}{\text{actual\_runtime}}
\end{equation}

We use two thresholds, $t_1$ and $t_2$, for the relative error, as explained below.

The confidence interval of the random forest method is estimated based on the prediction of each decision tree, $y_j, \forall j \in \{1,2,3,\cdots, J\}$. We take the $p^{\text{th} }$ and $(100-p)^{\text{th} }$ percentile of the distribution of $y_j$. If the final prediction $y$ is within this range, we consider the prediction to have low variance, since the predictions from all trees are consistent. 
%For a new job instance, by using the appropriate percentiles of the distribution of $y_j$, we can quantify the confidence interval that the response value will fall with a given probability.

\noindent We define three confidence levels as follows:
\begin{description}
	\item[High] The prediction $y$ is within the range of $p^{\text{th} }$ and $(100-p)^{\text{th} }$ percentile of $y_j$, and the relative error is lower than threshold $t_1$;
	\item[Medium] The relative error is between $t_1$ and $t_2$;
	\item[Low] Other scenarios. 
\end{description}

Parameters $p$, $t_1$ and $t_2$ are tuned as hyper parameters using validation data.
% \KK{We might define threshold if we have time.}
%We empirically define a threshold for these two factors. 
%When both factors' value is below the corresponding threshold, the confidence is considered high.
High confidence means our model can reliably predict the slowdown reasons. Low confidence means the reasons are likely to fall outside the metrics we used. In the API, the user will be presented the level of confidence. A low confidence indicates more investigation will be needed. However, even in this case, as the model has examined many metrics, the DRI can focus their investigation in other areas.

\section{Data Preparation and Feature Engineering}\label{sec:data collection}

% In this section, we first discuss feature engineering in \sysname, and then describe the data collection and pre-processing.

% Data preprocessing was conducted in Python.  A similar data preprocessing pipeline is used for training and prediction data. 
% Lacking labeled data, in this paper, we develop a job run-time prediction model, i.e. a regression model, instead. Afterwards, the increase in job run-time is attributed to the feature(s) of the regression model, formulating the reasoning of slowness. Thus, a direct link between the input features and the slowness reason was developed. And a carefully selected set of features that relates to various slowness reasons is required.  

\subsection{Data Preparation}\label{dataprep}

% Having chosen the right features, we now turn to the data collection process, which comes with its own challenges.
In our clusters, we keep hundreds of metrics for monitoring, reporting, and troubleshooting purposes, which result in petabytes of logs and metrics per day. 
Moreover, the features we are interested in are scattered both \emph{physically} (in different files across our hundreds of thousands of machines) and \emph{logically} (we need to process and combine different files to generate features).
To perform the required data preparation and extract features out of the data, we use Scope~\cite{scope}, which provides a SQL-like language and can support our scale.
% The resulting features are downloaded in a central location to be used by our models.

Feature extraction occurs both during the offline training phase and the online prediction (see Figure~\ref{fig:frm}).
Given that data freshness is not an issue for training (we do not need data of the current day), we use data that becomes available daily in our clusters and includes years-worth of historical data. For prediction, we need to collect features only for a single job (or a group of jobs), but latest data is required, as a user might want to debug their job that just finished. Thus, we use different data sources that allow us to access data within minutes from when they are produced (but that do not allow access to historical data, so cannot be used for training).

%--------------------------

\subsection{Feature Engineering and Selection}\label{exploreselection}

%The features are data from system logs that can help to detect and diagnose why a job is running slowly. Much relevant information is tracked in system logs about individual jobs. 

In collaboration with domain experts, we selected a subset of the features we collect to train our models in \sysname, based on what could potentially impact the runtime of a job.
As already discussed, a job can experience a slowdown compared to its previous occurrences, due to either user-induced or system-induced reasons. User-induced reasons can be captured by metrics collected at the job-level, whereas metrics related to system-induced reasons can be split to either machine-level or cluster-level, as detailed below.

\begin{description}
	\item[Job-level] These are metrics collected for each job. \sysname currently uses approximately 15 such features, including data read within and across racks, data written, data skewness metrics, job priority, execution DAG features (e.g., number of stages and tasks), and user information.
	\item[Machine-level] These are metrics collected at the machines used during the execution of a job, such as CPU load, allocation delays, I/O reads/writes. Due to the challenges of collecting such features and correlating them with each job, in production \sysname currently uses only a few of them, but we are working on adding more.
	\item[Cluster-level] These relate to the cluster environment when a job was executed. Examples of the ones we use are job queuing times, number of failed and revoked vertices, and execution environment version.
\end{description}

\mypara{Challenges} \label{challenges}
One of the biggest challenges with feature engineering is the correlation between features, which often results in high variance in the model prediction. While certain highly correlated features ($>0.95$) were removed, feature data was preserved to the greatest extent, because correlated features may indicate different problems with a job. For instance, input size and input size per task have a correlation of $0.9$. However, the former might indicate that the slowness reason was more data; the latter might indicate data skew.
% Input size might indicate that slowness was caused by more data, while input size per task might indicate a different problem, such as data skew. 
Fortunately, \sysname's tree-based models have an innate feature of being robust to correlated features.

\section{Tracking and Deployment}\label{sec:deployment}

\mypara{Model Tracking}
Machine learning is an iterative process, and reproducibility and versioning are crucial to productionalize machine learning models. To track model history in \sysname, we use MLflow~\cite{MLflow} in our Tracking Server (see Figure~\ref{fig:frm}), which is deployed as an Azure Linux VM. For each model, we track logging parameters, code versions, metrics, and model artifacts.
% MLflow's REST APIs publish performance metrics, hyper-parameters and artifacts (see Fig \ref{fig:frm}).

% \subsection{Deployment} \label{Model Deployment}
% The implemented architecture has two deployables. The Model Server is running as an Azure Container Instance, and \sysname Prediction Service is running as a flask web server. 
%We discuss their functionalities below.

\mypara{Model Serving}
In order to make available to end users a model that we trained and stored in the Tracking Server, we use one of MLflow's ``flavors''\footnote{An MLflow flavor is a convention that deployment tools use to understand the model.} to build an Azure Machine Learning image out of that model. Then we deploy this image as an Azure Container Instance, using the Azure ML Service's~\cite{AML} in the Model Server (see Figure~\ref{fig:frm}). 

The models that are deployed in the Model Server are made available to end users through a web application, which exposes a scoring API.
The web application runs on a flask web server~\cite{Flask}.
% that runs on a conda environment~\cite{Conda}.

% The models used for prediction are deployed as Azure Container Instances using the Azure Machine Learning Service~\cite{AML}. We use one of MLflow's ``flavors'' (which is a convention that deployment tools use to understand the model) to build an Azure ML image from one of the model artifacts that we have trained and made available on the Tracking Server.
%  % The Azure ML image is then deployed on an Azure Container Instance using the Azure ML Service's python SDK.

% The deployed container instance runs on a web server. It exposes a score API for predicting the run time and slowness reasons as shown in Fig \ref{fig:frm}.

% \subsubsection{Online Feature Building} \label{FeatureBuilding}

% \mypara{Web Application}
% The models deployed in the Model Server are made available to end users through a web application (see Figure~\ref{fig:frm}), which exposes a scoring API.

% The web application exposes a score API to the front-end user. The API takes a Job GUID and a Virtual Cluster (VC) name as inputs. It then collects the feature from the Online Feature Building component explained in \ref{online prediction}. A request and its associated features are sent to the Model Server VM which returns the predicted run time and slowness reasons to the user. 

% This service supports online prediction, which means predictions was generated instance by instance. The web application runs on a flask web server~\cite{Flask} running on a conda environment~\cite{Conda}.

\mypara{Model Monitoring} \label{monitoring}
It is critical to monitor the model performance and retrain models if they go stale. To ensure this, we have provisioned our pipeline to allow single click retraining. Decoupling data sources from the training pipeline helps to easily refresh our data, retrain, and deploy the model with minimal impact to end users. 

\section{Evaluation}\label{sec:experiment}

We now present our experimental evaluation for \sysname. In Section~\ref{sec:validation} we discuss \sysname's effectiveness in finding the actual causes of job slowdowns.
In Section~\ref{choosemodel} we compare different machine learning models for training, whereas Section~\ref{scale} studies the scalability of different models as the number of job templates increases.
In Section~\ref{sec:size} we compare the model performance when training our model with an increasing number of jobs per template.
In Section~\ref{modgeneralize} we show early evidence that \sysname can be applicable in different domains.

%The experiments were performed on Windows and Linux machines running a Conda environment in Python 3.7. The experiments are not computationally intensive and can be reproduced on any Windows machine with access to training data. 
We carried out our experiments on Windows using Python 3.7. We used a machine with eight 2.90 GHz processors and 64 GB RAM for the experiments in Sections~\ref{sec:validation} and~\ref{choosemodel}, and a high-memory Virtual Machine (VM) for the scalability experiments of Section~\ref{scale}.
%on a  with two 2.30 GHz processors and 432 GB RAM.

%\subsection{Datasets} 
The job and feature data is obtained from the \companyname production clusters, as described in Section~\ref{dataprep}. 
We use historical job data with different job templates, as described in Section~\ref{exploreselection}, over a period of three months.
%Each job template contains similar recurring jobs.
% For our analysis we extracted templates of six high value SLO jobs as training data.

\begin{figure}
	\centering
	\includegraphics[width=0.8\columnwidth]{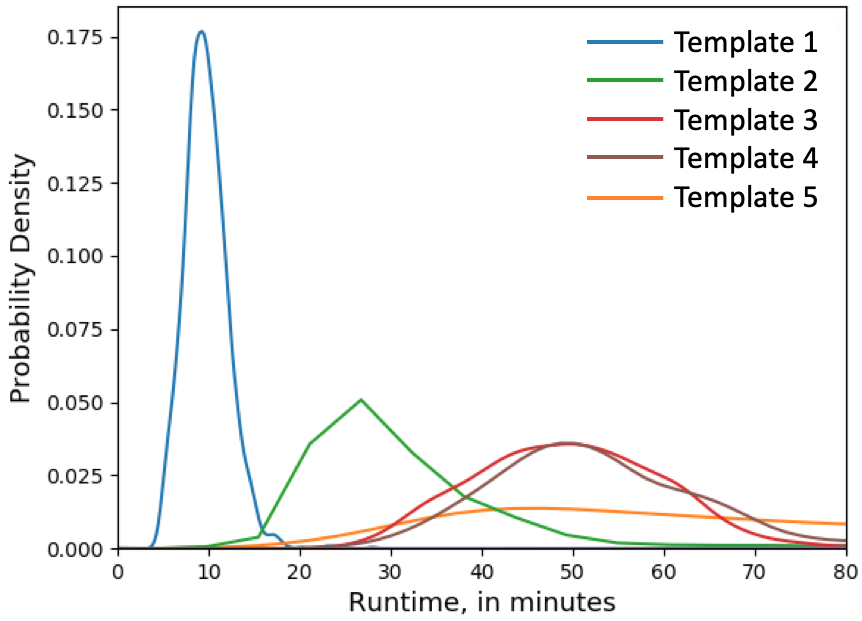}
	\caption{Runtime distribution for the jobs of different templates.} \label{fig:template-runtime}
\end{figure}

\subsection{Validation Results}\label{sec:validation}

Working with domain experts at \companyname, we picked a set of job templates that are considered important for our production clusters (SLO critical), and trained \sysname based on those.
Note that the runtime distribution of the jobs of different templates varies significantly, which poses extra challenges for the runtime prediction based on machine learning models.
Figure~\ref{fig:template-runtime} shows the runtime distributions for five of the templates that we used.

From these templates, we then randomly picked seven jobs that experienced slowdowns (five from these templates and two from different templates), and compared the causes for slowdowns that were identified by the experts with those suggested by \sysname.
For these jobs, Table~\ref{validation} shows the reasons identified by the experts and \sysname (with their ranking), \sysname's confidence level, and if the job belonged to one of the templates used for training the model (in-t). We use $R_x$ to denote the reason that \sysname predicted for a job's slowdown with rank $x$. For readability, we show only the reasons that were common between \sysname and experts and use $R_x$ variables for the rest.
%We list the important feature and associated reason which is consistent with human generated reason. We also show the rankings with confidence level, the impact on job runtime (IJR) in minutes and if the job template is in the training data (IF\_J). 
%The IJR shows how many minutes that this feature contributes to the runtime.
% Engineers focus on discovering root causes of job slowdown. \sysname predicts a set of reasons ranked with positive contribution to the slowness. If a feature contributes negatively, it is a potential factor to speed up the job.

\begin{table*}
	\begin{center}
		\caption{Result subset validated by engineers.}\label{validation}
		\begin{tabular}{cp{5cm}p{3.5cm}p{2cm}c}
			\toprule
			\bf Job & \bf \sysname's Predicted List of\newline Ranked Reasons & \bf Engineer Validated\newline Reason & \bf Confidence Level & \bf in-t \\ 
			\midrule
			1 & [Input Size, $R_2$, $R_3$] & Input size & High & Yes\\ 
			2 & [$R_1$, $R_2$, $R_3$, Revocation, $R_5$] & Revocation & Medium & Yes\\ 
			3 & [$R_1$, $R_2$, $R_3$, $R_4$, $R_5$, $R_6$] & Framework issue & Low & Yes\\ 
			4 & [$R_1$, $R_2$, $R_3$, $R_4$, High compute hours] & High compute hours & Medium & Yes\\ 
			5 & [Time skew, $R_2$, $R_3$, $R_4$] & Time skew & High & Yes\\ 
			6 & [High compute hours, $R_2$, $R_3$] & High compute hours & High & No\\ 
			% 1590 & [High compute hours, $R_2$] & High compute hours & High & No\\ 
			7 & [$R_1$, Usable machine count, $R_3$, $R_4$] & Usable machine count & High & No\\ 
			8 & [High compute hours, $R_2$] & High compute hours & High & No\\
			% DataWrite & Input size & +24.86m & 1 & High & Input Size & Yes\\ \hline
			% MaxVertexExecutionTime & Time skew & +3.25m & 1 & High & Time skew & Yes\\ \hline
			% RevocationCount & Revocation & +5.97m & 4 & Medium & Revocation & Yes\\ \hline
			% BonusPnHoursRatio & Yarn or cluster issue & +5.05m & 3 & Medium & Yarn or cluster issue & Yes\\ \hline
			% QueueTime & Cluster issue & -0.05m & - & Low & cluster issue & Yes\\ \hline
			% TotalPnHours & PN hours & +2.05m & 5 & Medium & High PN hours & Yes\\ \hline
			
			% TotalPnHours & PN hours & +16m & 1 & High & High PN hours & No\\ \hline
			% TotalPnHours & PN hours & +15.61m & 1 & High & High PN hours & No\\ \hline
			% BonusPnHoursRatio & Usable machine count & +2.77m & 2 & High & Usable machine count & No\\ \hline
			\bottomrule
		\end{tabular}
	\end{center}
\end{table*}

The results in Table \ref{validation} show that the reasons generated by \sysname are highly correlated with the reasons manually validated by our domain expert engineers. % with positive IJR and high confidence level. 
For job 1, the top predicted reason is the same as the manually validated reason with high confidence. For jobs 2 and 4, our system predicted the validated reason in the top 5 slowdown reasons, which is consistent with the confidence level medium. For job 3, our model does not identify the same reasons as the experts, also consistent with low confidence.
%The Engineer validated reason does not show in the predicted list of reasons by \sysname, because the feature associated with DRI validated reason negatively contributes to the job slowdown.
Adding more features as planned (e.g., additional machine-level features) will allow us to improve the model’s prediction capability and minimize such low-confidence cases. Jobs 6, 7 and 8 show the robustness of our model: although \sysname was not trained using these job templates, it can still find the correct reasons with high confidence by using knowledge gathered by other similar job templates.
Importantly, we observed no misleading predictions, i.e., there were no cases where \sysname predicted wrong slowdown reasons with high confidence. This means that even predictions with low confidence can be useful in ruling out the currently used features from the investigation.

\subsection{Picking the Right Model}\label{choosemodel}

As discussed in Section \ref{sec:reasoning}, we experimented with various categories of models for the job runtime prediction, including Linear Regression (LR), Random Forest (RF), Gradient Boosted Trees (GBT) and Deep Neural Networks (DNN) with two hidden layers without hyper parameter tuning. For each of these categories, we consider both a global and per-template models.
%In our case we generate six unique models based on jobs from six job templates.
%One drawback of the per-template approach is that if there are new jobs with unknown job templates, we will need to train a new model to handle them effectively.

%We carry out experiments over six job templates data in Table \ref{tab:model}. 
We use Mean Absolute Ratio Error (MARE) as a metric to evaluate each model's accuracy. As the runtime distribution varies significantly across different job templates, we normalize the estimation error by the baseline runtime of each job template (see Section~\ref{sec:propstate}), calculating the average runtime per job template in the training data:
\begin{equation}\label{eq:mare}
\text{MARE} = \dfrac{1}{n}\sum_{i=1}^n |\dfrac{y_{\text{test}_i}-\hat{y}_{\text{test}_i}}{y_{\text{test}_i}^{\beta}}|
\end{equation}

% \begin{table}[ht]
% \caption{MARE scores on runtime prediction by Linear Regression (LR), Random Forest (RF), Gradient Boosting (GB) and Deep neural networks (DNN) with global and single models. Lower MER is better.}\label{model}
%  \begin{center}
% \begin{tabular}{l|c|c|c|c}
% \hline \bf  & \bf LR & \bf RF & \bf GB & \bf DNN \\ \hline
% Per-Template Model & 0.186 & \bf 0.116 & 0.124 & 0.146 \\
% \hline
% Global Model & 0.235 & \bf 0.121 & 0.277 & 0.353 \\
%  \hline
% \end{tabular}
% \end{center}
% \end{table}

\noindent where $n$ is the number of jobs in the testing data, %$\bar{y}_{train,m}$ is the average runtime of the corresponding job template $m$ in the training data. 
$y_{\text{test}_i}$, $\hat{y}_{\text{test}_i}$, and $y_{\text{test}_i}^{\beta}$ are the predicted, actual, and baseline runtime from testing data, respectively. 
% We use MARE to evaluate how good our predicted results compared to the actual values, and the deviation from the mean. %In this project, we predict runtime compared with the actual runtime for different jobs.

\begin{figure*}[h]
	\centering
	\begin{subfigure}[b]{0.4\textwidth}
		\centering
		\includegraphics[width=\textwidth]{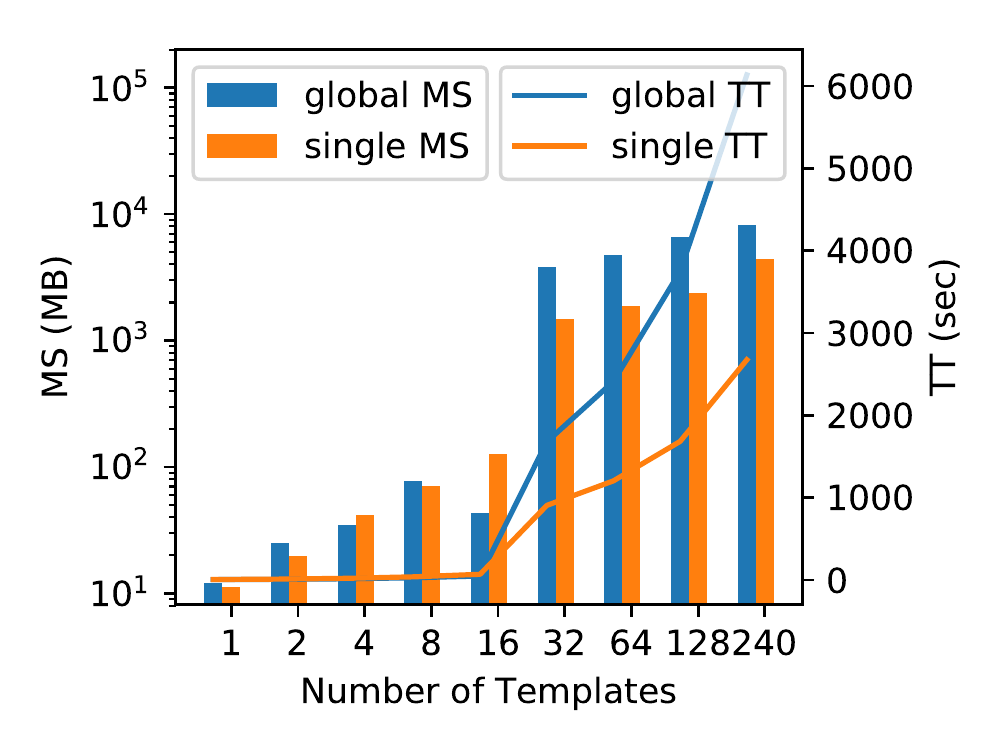}
		\caption{MS and TT}
	\end{subfigure}%
	\begin{subfigure}[b]{0.4\textwidth}
		\centering
		\includegraphics[width=\textwidth]{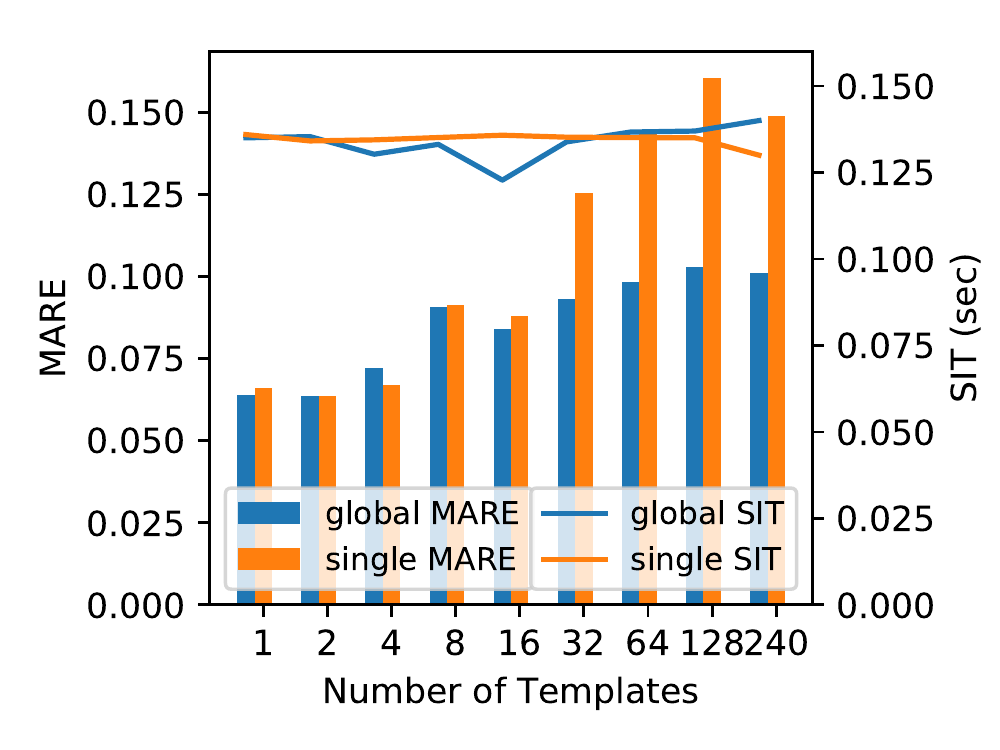}
		\caption{MARE and SIT}
	\end{subfigure}%
	\caption{Scalability of global vs. per-template models}\label{fig:more_temp}
\end{figure*}

Table \ref{tab:model} shows the results of MARE scores for the four model categories. Random forest performs best in terms of accuracy, both for the global and per-template model. Given its high accuracy and interpretability (as discussed in Section~\ref{sec:model}), RF is the approach we use in our production \sysname deployments. Moreover, we observe that the global model tracks closely the performance of the per-template model, while allowing to reason about jobs that we have not sufficiently encountered previously. In the next section, we demonstrate that the global model scales much better than the per-template models with an increasing number of templates. Hence, we use the global model in production. 
%We see that for all the models the per-template model's achieves slightly better performance than the equivalent global model and random forest achieves the best performance.  
%We choose to use a single global model for our final product, despite obtaining slightly higher performance with per-template models (Table \ref{tab:model}), in order to maximize extensibility to future unknown jobs and unknown job templates.

\begin{table}[t!]
	\centering
	\caption{MARE scores on runtime prediction by LR, RF, GBT, DNN with global and per-template models. Lower MARE is better.} \label{tab:model}
	\begin{tabular}{p{2cm}p{1cm}p{1cm}p{1cm}p{1cm}}
		\toprule
		\bf  & \bf LR & \bf RF & \bf GBT & \bf DNN  \\
		\midrule
		\ Per-Template Model & 0.186 & \bf 0.116 & 0.124 & 0.146 \\
		\ Global Model & 0.235 & \bf 0.121 & 0.277 & 0.353 \\
		\bottomrule
		
	\end{tabular}
\end{table}

\begin{figure}[t!]
	\centering
	\includegraphics[width=0.9\columnwidth]{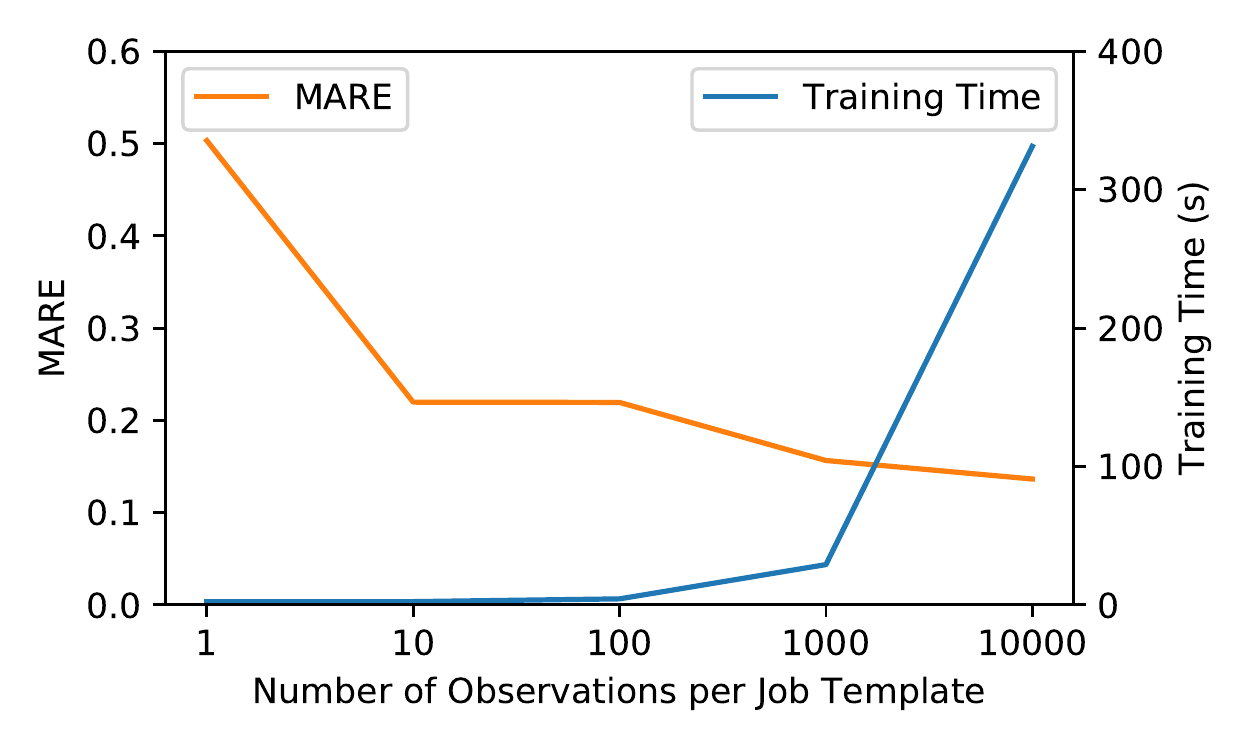}
	\caption{Model performance with different training sample sizes} \label{fig:sizeb}
\end{figure}

\subsection{Scalability of Global vs. Per-template Models}\label{scale}
We assess the scalability of the global and per-template models by training them with an increasing number of job templates, as shown in Figure~\ref{fig:more_temp}. We added job templates incrementally to the training set and repeated each experiment 10 times. We use 5-fold cross validation for hyper parameter tuning with random grid search. 
With more job templates, the training time (TT) and the model size (MS) for both global and per-template models increased. 
We observe that the MARE for global model is better than per-template models when training with a larger number of job templates. 
This can be attributed for the most part to the larger sample size for the global model, as a result of the unification of the training set. 
In contrast, when training the per-template models with 240 job templates, many templates only have a small number of samples and the MARE is high. 
%The model is applicable for a large number of job templates with similar MARE (see~Fig \ref{fig:more_temp}), though the training time (TT) increased. 
%We also report the prediction time on around 150k jobs, i.e. batch inference time (BIT) and on single job, i.e. single inference time (SIT). 
We also report the prediction time on a single job, i.e. the single inference time (SIT). 
The SIT didn't increase with the model size, which is important to deliver a real-time experience to \sysname's users.
Overall, the above experiments demonstrate the scalability of the global model for cloud-scale training. 
At the other extreme, a template-specific model suffers from lack of training data and the ability to generalize to new (unseen) templates.
As part of our future work, we plan to cluster job templates using unsupervised machine learning methods and train ``semi-global'' models that take into account multiple job templates that share similar characteristics to strike a balance between the two approaches.
% with respect to multiple job templates that share similar characteristics.
%A model trained with multiple job templates is more applicable. 

\subsection{Varying Size of Training Data} \label{sec:size}

In order to determine the impact of the number of jobs per template on our model performance, 
we retrain our global model assuming that we only have a limited number of observations for each job template. 
In particular, we train our model with $n$ observations per job template, where $n \in \{1, 10, 100, 1000,10000\}$. 
The MARE and the training time are reported in Figure~\ref{fig:sizeb}. 
We used 22 job templates and, for each run of the experiment, a random sample of $n$ observations were selected as the training data. 
When $n\geq 10$, the MARE dropped below 20\%, and the prediction accuracy continues to improve for larger sample sizes, although less significantly. 
Note that the training time increased exponentially to the sample size.

\subsection{Model Generalization} \label{modgeneralize}

The baseline approach for interpretation described in this paper allows job runtime prediction results to be interpreted and compared to a set of similar jobs. Data in the real world frequently looks similar to our dataset: Gaussian mixture distributions of a target variable are commonly encountered.   

This section presents an example of employing \sysname on another dataset in another field.  A classic dataset from statistics is the ``Auto'' dataset of gas mileage, which is represented well by a linear superposition of three Gaussians by region of origin: American, European and Japanese. Manufacturers might be interested in understanding what factors drive higher gas mileage in American cars relative to other American cars. Here gas mileage is the equivalent to job runtime. %Figure~\ref{fig:mpg} shows 
% The distribution of gas mileage and follows a similar pattern to runtime distribution described in Section \ref{exploreselection}.

%\begin{figure} [ht]
%\centering
%\includegraphics[scale=.40]{mpg.png}
%\caption{Gas mileage by origin shows underlying Gaussian distributions} \label{fig:mpg}
%\end{figure}

% As shown in Figure~\ref{fig:autofeatures},
% After investigation, features are similarly mixture distributions. A researcher may be more interested in the impact of auto weight compared to cars of similar origin.  For example, if a Japanese car is predicted to have lower gas mileage due to weight relative to other Japanese cars. 
% \begin{figure} [h]
% \includegraphics[scale=.25]{weight.png}
% \includegraphics[scale=.25]{horsepower.png}
% \caption{Features exhibit mixture distribution} \label{fig:autofeatures}
% \end{figure}

% \begin{table}[ht]
%   \centering
%   \begin{tabular}{p{1.65cm}p{0.85cm}p{0.65cm}p{0.9cm}p{1.1cm}p{1.1cm}}
%     \toprule
%     Feature  & Ford Granada & All Cars  &  American baseline & Overall contribution (in mpg) & Delta FC (in mpg)  \\
%     \midrule
%     \ Weight  & 3060 & 2978 &   3224 & -0.12 & 0.7 \\
%     \ Year & 81 & 76 & 75 & 2.13 & 2.07 \\
%     \ Horsepower    & 88 & 104 & 106 & 1.11 & 1.07   \\
%     \ Acceleration     & 17.1 & 15.5 & 16.1 & -0.19 & -0.19  \\    
%     \ Displacement  & 200 & 194 & 221  &-0.26 & 0.04   \\
%     \ Cylinders & 6  & 5.5 & 5.8 & -3.53 & -1.4\\  
%     \bottomrule

%   \end{tabular}
%   \caption{Ford Granada} \label{tab:granada}
% \end{table}

\begin{table}[t!]
	\centering
	\caption{Contribution of each feature to the high gas mileage of American-made Ford Granada compared to other American cars.} \label{tab:granada}
	\scalebox{0.9}{
		\begin{tabular}{p{2cm}p{1cm}p{1cm}p{1cm}p{1cm}}
			\toprule
			Feature  & Ford \linebreak Granada & Mean & Baseline & Delta FC (in mpg)  \\
			\midrule
			\ Year & 81 & 76 & 75 & 2.07 \\
			\ Horsepower    & 88 & 104 & 106 & 1.07 \\
			\ Weight  & 3060 & 2978 &   3224 & 0.7 \\
			\ Displacement  & 200 & 194 & 221 & 0.04   \\
			\ Acceleration     & 17.1 & 15.5 & 16.1 & -0.19  \\    
			\ Cylinders & 6  & 5.5 & 5.8 & -1.4\\  
			\bottomrule
			
		\end{tabular}
	}
\end{table}

Table~\ref{tab:granada} summarizes the delta contribution of each feature (FC) to the high gas mileage of American-made Ford Granada compared to other American cars based on our anomaly reasoning algorithm, described in Section \ref{sec:reasoning}.
%Features included weight, horsepower, acceleration, displacement, year and cylinders. 
We observe that ``Year" and ``Horsepower" contribute the most to high gas mileage, while ``Weight" and ``Displacement" make marginal contributions. ``Acceleration" and ``Cylinders" contribute to low gas mileage.

\vspace{-1mm}
\section{Related Work}\label{sec:related}
\vspace{-2mm}

Anomaly detection \cite{Chandola2009AnomalyDA} refers to the problem of finding patterns in data that do not conform to expected behavior. In contrast, \emph{anomaly reasoning}, which is the purpose of this work, encompasses recognizing, interpreting, and reacting
to unfamiliar objects or familiar objects appearing in unexpected contexts. 
% Building an anomaly reasoning system has been viewed by many industry experts as the cornerstone to reduce human labor~\cite{jaw2002anomaly},
% as it would improve speed and quality of human learning, minimize failures, and ensure robustness.

Anomaly reasoning is particularly important for large-scale systems, as it is not possible to manually track all machines and applications at scale. Below we discuss the main categories of works that are related to \sysname.

\mypara{Interpretable models}
% \YZ{I feel this part goes to section 3 better?}
Building an effective anomaly system requires both interpretable models and reasoning algorithms. 
Several efforts have focused on interpreting the results of machine learning models. Their goal is to provide proper explanation about how or why the algorithm produces a specific prediction and to identify interactions between features and estimation results~\cite{dovsilovic2018explainable,gunning2017explainable,samek2017explainable}.
In \sysname, we use an interpretable random forest model to rank a job's slowdown reasons given the contribution of various features. Similar methods to interpret a tree model can be generalized to boosting algorithms~\cite{welling2016forest}.
% Such a system would improve the speed and quality of human learning, minimize failures, and ensure robustness of large-scale systems.

%To this end, there have been several efforts to detect anomalies and respond rapidly to customer issues, based on statistical and machine learning methods. 

\mypara{Anomaly reasoning with labeled data}
Existing related works have used detection methods such as classification and clustering to perform analysis of anomalies in cloud computing. 
A detailed survey of those work can be seen in~\cite{agrawal2015survey,modi2013survey}. 
For example, a fault detection and isolation system based on k-nearest neighbor has been proposed to rank machines in order of their anomalous behavior~\cite{bhaduri2011detecting}. 
%The Fault Detection in Cloud Systems (FDCS) framework utilizes a distance-based outlier ranking function to detect anomaly in an adaptive way. Based on the outlier score, the fault isolation can be done by examine the contribution of each variable towards the outlier score based on euclidean distance.   
Other works have used a hybrid of SVM classification and k-medroids clustering to detect intrusions of the network~\cite{chitrakar2012anomaly}. An anomaly-based clustering method has also been suggested to detect failures in general production systems~\cite{duan2009fa}.

The downside of these approaches is that they require labeled data.
% , e.g., data from existing incidents that associate jobs with their slowdown causes in our case.
Such data is hard to acquire in many settings, including ours. In the context of an infrastructure that has been operating for many years,
labeling data requires infrastructure support and, most importantly, training a large number of engineers to add labels when resolving anomalies. 
Moreover, it is almost impossible to perform labeling for the years-worth of historical data.

% However, to identify the reason for slowdowns, these methods require labeled data, e.g.,  
% Such labeled training data in production cloud systems are extremely hard to obtain and can also be erroneous.

\mypara{Anomaly reasoning with unlabeled data}
A few approaches have considered unlabeled data, but focus either on time-series analysis or are restricted to machine or VM metrics~\cite{cherkasova2009automated,cohen2004correlating,dean2012ubl,gu2009online,tan2010adaptive,tan2012prepare,zhang2013intelligent}.
% \KK{This does time series. Features are continuous and are time-correlated. Analysis on-demand, not continuously.}
For instance, an anomaly detection and reasoning system has been proposed to detect security problems during VM Live Migration~\cite{zhang2013intelligent}.
This work is based on time series data related to resource utilization statistics, e.g., file read/write, system call, CPU usage. %The algorithm consists of two steps: adaptive Dimension Reasoning-Local Outlier Factors (DF-LOF) to detect collective anomalies and Symbolic Aggregate
%Approximation (SAX) algorithm to determine if applications operate normally after the migration. 
%The method successfully identified anomalies of CPU emulation, memory leakage, etc.
% \YZ{time series}
PREdictive Performance Anomaly pREvention (PREPARE)~\cite{tan2012prepare} predicts performance anomalies using a 2-dependent Markov model and a classifier based on system-level metrics, such as CPU, memory, network traffic. 
%A  tree-augmented Bayesian networks (TAN) model is used for anomaly causes. 
%Similar to \sysname, the TAN model estimates the contribution of each attribute to the anomaly and provides information about which system-level metrics to look into. 
%However, the method only predicts an anomaly that has been observed before and relies on time-dependency.
% \YZ{time series}
% \KK{check further -- metrics are given just as clues for slowdowns. They use neurons. They look at machine-level data in a time-series fashion. They have a very small number of features. For them it matters to find anomalies in data series... Not across occurrences of a job. Very few results on root cause predictions.}
Another related work developed an Unsupervised Behavior Learning (UBL) system to capture the anomalies and infer their causes~\cite{dean2012ubl}. To circumvent the need for labeling data, Self Organizing Map (SOM) has been suggested to model the system behavior, and deviations are used for the anomaly detection~\cite{kohonen2012self}. 
% The method is less computational intensive than the k-nearest neighbor. 

Similar to \sysname, those methods estimate the contribution of each attribute to the anomaly and provides information about which system-level metrics to look into. 
However, they require time-dependent series of data to capture the anomalies. In contrast, we focus on job instances that span several hundreds of machines but only during the lifetime of the job. Thus, our features are neither time series nor machine-centric (although we do employ some system-level data to examine the system's impact on a particular job's execution).
%By comparing the Euclidean distance form the normal neuron to the anomalous neuron in SOM, the metrics that differ the most is the top faulty one. Implemented in a real world distributed systems, the results show that UBL can achieve high prediction accuracy. However, this method also require time-dependent time series data to capture the anomalies.

Other works in anomaly reasoning aim to pinpoint the faulty components of a system by tracing the system's activities~\cite{aguilera2003performance,mi2012performance,mi2011magnifier,nguyen2011pal}. The methods rely more on the estimation of the time series' change point and the propagation pattern or the execution graph. However, those methods require significant domain knowledge and are hard to generalize.

%.
% It is desirable to develop a robust diagnosis and forecasting mechanism to automatically detect anomaly and response rapidly to customers~\cite{cherkasova2009automated,zhang2013intelligent,bhaduri2011detecting,cohen2004correlating,duan2009fa,tan2010adaptive,gu2009online}. Anomaly reasoning for cloud computing system has been studied based on statistical and machine learning methods.
To the best of our knowledge, \sysname is the first anomaly reasoning system to be deployed at this scale in production to identify the causes of job slowdowns in analytics clusters. Unlike existing approaches, it follows a job-centric approach and does not rely neither on labeled data nor on time series analysis.
% on distributed database in a real production environment (with thousands of nodes) to be described in the literature. 

%To the best of our knowledge, our system is the first anomaly reasoning system on distributed database in a real production environment (with thousands of nodes) to be described in the literature. 
\vspace{-1mm}
\section{Conclusion \& Future Work}\label{sec:conclusion}
\vspace{-2mm}

We presented \sysname, a system that we built and have deployed in production to detect the causes of job slowdowns in \companyname's big data analytics clusters, consisting of hundreds of thousands of machines.
\sysname does not require labeled data to perform anomaly reasoning. Instead, it uses an interpretable machine learning model to predict the runtime of a job that has experienced a slowdown. Using this model, we can determine the contribution of each feature in the deviation of the job's runtime compared to previous normal executions of the job (or of jobs with similar characteristics).

Our evaluation results using historical incidents showed that \sysname discovers the same slowdown reasons that were detected by domain expert engineers. We also compared various categories of models and showed that a global (i.e., trained over all jobs) random forest model strikes a good balance between accuracy, training time, model size, and generalization capabilities. 
% Finally, we showed initial evidence that \sysname could be applicable in different domains.

\mypara{Towards data-driven decisions} \sysname is part of our bigger vision towards employing data-driven decisions to optimize various aspects of our systems.
Taking \sysname's capabilities a step further, knowing the job slowdown reasons allows us to automatically tune the system to avoid such slowdowns in the future.
This may include both system parameters, such as dynamically setting the number of running tasks per machine, and application parameters, such as the degree of parallelism for each stage of a job.
Moreover, such parameter autotuning does not have to be constrained to job slowdowns---we can use it to automatically and dynamically set various parameters in our systems to improve their performance.

Furthermore, although \sysname currently targets our internal analytics clusters, the above techniques can be applied to other environments, such as various public Azure services, including the Azure SQL and HDInsight offerings. Similar data-driven decisions are increasingly applied in various companies~\cite{uberds,linkedinds}.

% our model on a portion of the historical dataset and show that the output of our model is interpretable by engineers and achieves good performance in discovering the reasons for job slowdowns that are consistent with human-labeled reasons. In addition, we illustrate the generalization of the model by applying our model on the 'Auto' dataset. 

% Our partners in the Big Data Fundamentals team are currently migrating \sysname from a corporate machine to production one by end of March 2019 and extending the pipeline to different Azure products having similar problems.
% When the training data size increase causes scalability issues, \sysname is also planned to be migrated to a distributed system.
% Finally, plans are also underway to automate the model validation process from manual validation by DRIs.

% consisting of an end-to-end pipeline for the online prediction of the reasons for the slowing down of Cosmos jobs under specific real-time requirements and an interpretable ML model. 

\balance

{\small

\bibliographystyle{abbrv}
\bibliography{soccbib}

\begin{thebibliography}{10}

\bibitem{agrawal2015survey}
S.~Agrawal and J.~Agrawal.
\newblock Survey on anomaly detection using data mining techniques.
\newblock {\em Procedia Computer Science}, 60:708--713, 2015.

\bibitem{aguilera2003performance}
M.~K. Aguilera, J.~C. Mogul, J.~L. Wiener, P.~Reynolds, and A.~Muthitacharoen.
\newblock Performance debugging for distributed systems of black boxes.
\newblock In {\em ACM SIGOPS Operating Systems Review}, volume~37, pages
  74--89. ACM, 2003.

\bibitem{bhaduri2011detecting}
K.~Bhaduri, K.~Das, and B.~L. Matthews.
\newblock Detecting abnormal machine characteristics in cloud infrastructures.
\newblock In {\em 2011 IEEE 11th International Conference on Data Mining
  Workshops}, pages 137--144. IEEE, 2011.

\bibitem{Chandola2009AnomalyDA}
V.~Chandola, A.~Banerjee, and V.~Kumar.
\newblock Anomaly detection: A survey.
\newblock {\em ACM Comput. Surv.}, 41:15:1--15:58, 2009.

\bibitem{cherkasova2009automated}
L.~Cherkasova, K.~Ozonat, N.~Mi, J.~Symons, and E.~Smirni.
\newblock Automated anomaly detection and performance modeling of enterprise
  applications.
\newblock {\em ACM Transactions on Computer Systems (TOCS)}, 27(3):6, 2009.

\bibitem{chitrakar2012anomaly}
R.~Chitrakar and C.~Huang.
\newblock Anomaly based intrusion detection using hybrid learning approach of
  combining k-medoids clustering and naive bayes classification.
\newblock In {\em 2012 8th International Conference on Wireless Communications,
  Networking and Mobile Computing}, pages 1--5. IEEE, 2012.

\bibitem{owl}
A.~Chung, C.~Curino, S.~Krishnan, K.~Karanasos, G.~Ganger, and P.~Garefalakis.
\newblock {Peering through the dark: an Owl's view of inter-job dependencies
  and jobs' impact in shared clusters}.
\newblock In {\em {SIGMOD}}, 2019.

\bibitem{cohen2004correlating}
I.~Cohen, J.~S. Chase, M.~Goldszmidt, T.~Kelly, and J.~Symons.
\newblock Correlating instrumentation data to system states: A building block
  for automated diagnosis and control.
\newblock In {\em OSDI}, volume~4, pages 16--16, 2004.

\bibitem{hydra}
C.~Curino, S.~Krishnan, K.~Karanasos, S.~Rao, G.~M. Fumarola, B.~Huang,
  K.~Chaliparambil, A.~Suresh, Y.~Chen, S.~Heddaya, R.~Burd, S.~Sakalanaga,
  C.~Douglas, B.~Ramsey, and R.~Ramakrishnan.
\newblock Hydra: a federated resource manager for data-center scale analytics.
\newblock In {\em {NSDI}}, 2019.

\bibitem{dean2012ubl}
D.~J. Dean, H.~Nguyen, and X.~Gu.
\newblock Ubl: Unsupervised behavior learning for predicting performance
  anomalies in virtualized cloud systems.
\newblock In {\em Proceedings of the 9th international conference on Autonomic
  computing}, pages 191--200. ACM, 2012.

\bibitem{dovsilovic2018explainable}
F.~K. Do{\v{s}}ilovi{\'c}, M.~Br{\v{c}}i{\'c}, and N.~Hlupi{\'c}.
\newblock Explainable artificial intelligence: A survey.
\newblock In {\em 2018 41st International convention on information and
  communication technology, electronics and microelectronics (MIPRO)}, pages
  0210--0215. IEEE, 2018.

\bibitem{duan2009fa}
S.~Duan, S.~Babu, and K.~Munagala.
\newblock Fa: A system for automating failure diagnosis.
\newblock In {\em 2009 IEEE 25th International Conference on Data Engineering},
  pages 1012--1023. IEEE, 2009.

\bibitem{Flask}
Flask.
\newblock Flask - a python microframework.
\newblock \url{http://flask.pocoo.org/}, 2019.

\bibitem{gu2009online}
X.~Gu and H.~Wang.
\newblock Online anomaly prediction for robust cluster systems.
\newblock In {\em 2009 IEEE 25th International Conference on Data Engineering},
  pages 1000--1011. IEEE, 2009.

\bibitem{gunning2017explainable}
D.~Gunning.
\newblock Explainable artificial intelligence (xai).
\newblock {\em Defense Advanced Research Projects Agency (DARPA), nd Web},
  2017.

\bibitem{morpheus}
S.~A. Jyothi, C.~Curino, I.~Menache, S.~M. Narayanamurthy, A.~Tumanov,
  J.~Yaniv, R.~Mavlyutov, I.~Goiri, S.~Krishnan, J.~Kulkarni, and S.~Rao.
\newblock {Morpheus: Towards Automated SLOs for Enterprise Clusters}.
\newblock In {\em {OSDI}}, 2016.

\bibitem{kohonen2012self}
T.~Kohonen.
\newblock {\em Self-organizing maps}, volume~30.
\newblock Springer Science \& Business Media, 2012.

\bibitem{linkedinds}
{An Introduction to AI at LinkedIn}.
\newblock
  \url{https://engineering.linkedin.com/blog/2018/10/an-introduction-to-ai-at-linkedin},
  2019.

\bibitem{Mein2006}
N.~Meinshausen.
\newblock Quantile regression forests.
\newblock {\em Journal of Machine Learning Research}, 7:983--999, 2006.

\bibitem{mi2012performance}
H.~Mi, H.~Wang, G.~Yin, H.~Cai, Q.~Zhou, and T.~Sun.
\newblock Performance problems diagnosis in cloud computing systems by mining
  request trace logs.
\newblock In {\em 2012 IEEE Network Operations and Management Symposium}, pages
  893--899. IEEE, 2012.

\bibitem{mi2011magnifier}
H.~Mi, H.~Wang, G.~Yin, H.~Cai, Q.~Zhou, T.~Sun, and Y.~Zhou.
\newblock Magnifier: Online detection of performance problems in large-scale
  cloud computing systems.
\newblock In {\em 2011 IEEE International Conference on Services Computing},
  pages 418--425. IEEE, 2011.

\bibitem{AML}
Microsoft.
\newblock Azure machine learning service - build, train, and deploy models from
  the cloud to the edge.
\newblock
  \url{https://azure.microsoft.com/en-us/services/machine-learning-service/},
  2019.

\bibitem{MLflow}
MLflow.
\newblock Mlflow - a platform for machine learning lifecycle.
\newblock \url{https://github.com/mlflow/mlflow}, 2019.

\bibitem{modi2013survey}
C.~Modi, D.~Patel, B.~Borisaniya, H.~Patel, A.~Patel, and M.~Rajarajan.
\newblock A survey of intrusion detection techniques in cloud.
\newblock {\em Journal of network and computer applications}, 36(1):42--57,
  2013.

\bibitem{nguyen2011pal}
H.~Nguyen, Y.~Tan, and X.~Gu.
\newblock Pal: Propagation-aware anomaly localization for cloud hosted
  distributed applications.
\newblock In {\em Managing Large-scale Systems via the Analysis of System Logs
  and the Application of Machine Learning Techniques}, page~1. ACM, 2011.

\bibitem{TreeInterpreter}
A.~Saabas.
\newblock Treeinterpreter.
\newblock \url{https://github.com/andosa/treeinterpreter}, 2018.

\bibitem{samek2017explainable}
W.~Samek, T.~Wiegand, and K.-R. M{\"u}ller.
\newblock Explainable artificial intelligence: Understanding, visualizing and
  interpreting deep learning models.
\newblock {\em arXiv preprint arXiv:1708.08296}, 2017.

\bibitem{tan2010adaptive}
Y.~Tan, X.~Gu, and H.~Wang.
\newblock Adaptive system anomaly prediction for large-scale hosting
  infrastructures.
\newblock In {\em Proceedings of the 29th ACM SIGACT-SIGOPS symposium on
  Principles of distributed computing}, pages 173--182. ACM, 2010.

\bibitem{tan2012prepare}
Y.~Tan, H.~Nguyen, Z.~Shen, X.~Gu, C.~Venkatramani, and D.~Rajan.
\newblock Prepare: Predictive performance anomaly prevention for virtualized
  cloud systems.
\newblock In {\em 2012 IEEE 32nd International Conference on Distributed
  Computing Systems}, pages 285--294. IEEE, 2012.

\bibitem{uberds}
{How Uber Organizes Around Machine Learning}.
\newblock \url{https://urlzs.com/J4Rk9}, 2019.

\bibitem{welling2016forest}
S.~H. Welling, H.~H. Refsgaard, P.~B. Brockhoff, and L.~H. Clemmensen.
\newblock Forest floor visualizations of random forests.
\newblock {\em arXiv preprint arXiv:1605.09196}, 2016.

\bibitem{zhang2013intelligent}
Q.~Zhang, Y.~Wu, T.~Huang, and Y.~Zhu.
\newblock An intelligent anomaly detection and reasoning scheme for vm live
  migration via cloud data mining.
\newblock In {\em 2013 IEEE 25th International Conference on Tools with
  Artificial Intelligence}, pages 412--419. IEEE, 2013.

\bibitem{scope}
J.~Zhou, N.~Bruno, M.-C. Wu, P.-{\AA}. Larson, R.~Chaiken, and D.~Shakib.
\newblock {SCOPE}: parallel databases meet {MapReduce}.
\newblock {\em VLDB J.}, 21(5):611--636, 2012.

\end{thebibliography}
}

\end{document}